\documentclass[letterpaper, 10 pt, conference]{ieeeconf}  %

\IEEEoverridecommandlockouts
\usepackage{booktabs}
\usepackage{cite}
\usepackage{cuted}
\usepackage{amsmath,amssymb,amsfonts}
\usepackage{algorithmic}
\usepackage{graphicx}
\usepackage{hyperref}
\usepackage[capitalise, nameinlink]{cleveref}
\usepackage{makecell}
\usepackage{multirow}
\usepackage{textcomp}
\usepackage{xcolor}
\usepackage{xspace}

\usepackage[normalem]{ulem}

\definecolor{mypastelgreen}{HTML}{5B8E7D}  
\definecolor{mypastelred}{HTML}{BC4B51}  
\definecolor{mypastelorange}{HTML}{F4A259}  
\definecolor{myearth}{HTML}{A8986B}

\definecolor{mygreen}{rgb}{0.12, 0.54, 0.30}
\definecolor{myred}{rgb}{0.78, 0.33, 0.37}
\hypersetup{
    colorlinks=true,
    linkcolor=mypastelgreen,
    filecolor=magenta,      
    urlcolor=myearth,
    citecolor=mypastelgreen,
}

\def\BibTeX{{\rm B\kern-.05em{\sc i\kern-.025em b}\kern-.08em
    T\kern-.1667em\lower.7ex\hbox{E}\kern-.125emX}}
\definecolor[named]{TODORed}{HTML}{A30B37}

\newcommand{\algname}{Fetch Interaction for Dynamics Observation (FIDO)\xspace}
\newcommand{\algabbr}{FIDO\xspace}
\usepackage{etoolbox}
\makeatletter
\patchcmd{\@maketitle}{\vskip0.25in}{\vskip0.0in}{}{\typeout{TITLEPATCH FAILED}}
\makeatother
\begin{document}

\title{\textbf{ FIDO the Robot Dog: \\
Learning Dynamics Models via In-the-Wild Fetch}
}

\author{Michelle Pan$^{1}$, Maximilian Du$^{1}$, Sriram Yenamandra$^{1}$, Ajay Sridhar$^{1}$, \\
Mark Zolotas$^{2}$, Dorsa Sadigh$^{1}$, Jennifer Grannen$^{1}$%
\thanks{This work was supported by the Toyota Research Institute (TRI) and the NSF (Award \#1941722). The authors are with $^{1}$Stanford University, and $^{2}$Toyota Research Institute. Correspondence to \texttt{mpan8@stanford.edu}.}
}

\maketitle

\begin{strip}
    \centering
    \vspace{-2.2cm}
    \includegraphics[width=\textwidth]{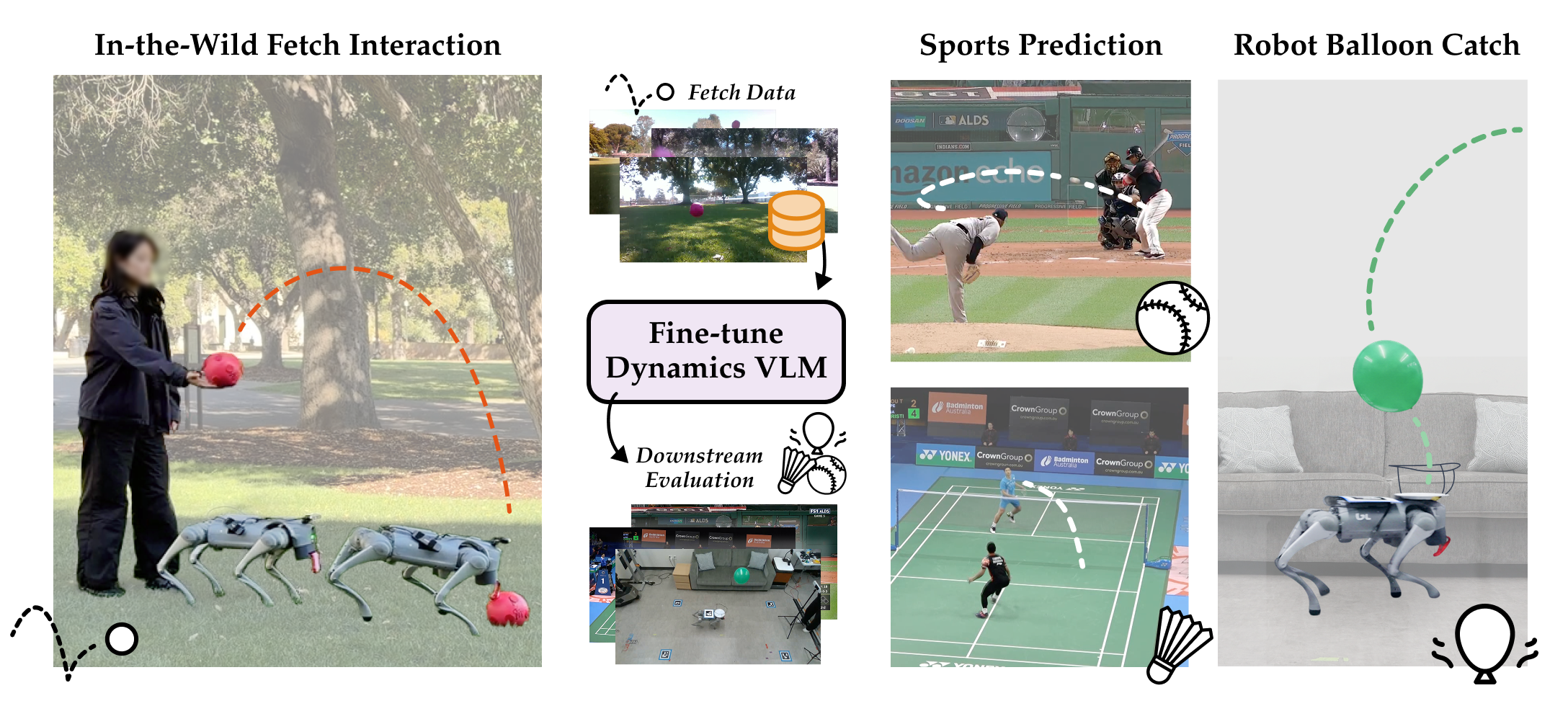}\par
    \vspace{-0.15cm}
    \refstepcounter{figure} %
    \begin{minipage}{1.0\textwidth}
    \footnotesize
    \setlength{\parskip}{0pt} %
    \setlength{\parindent}{0pt} %
    Fig.~\thefigure. 
    \textbf{FIDO: Fetch Interaction for Dynamics Observation}. Natural robot interactions can provide diverse dynamics data that improve models for downstream applications. We show this by designing a robot quadruped fetch game that collects crowdsourced ball trajectory data in a variety of environments [\textbf{Left}]. We fine-tune a Vision-Language Model (VLM) on the collected dynamics observations [\textbf{Center}] and demonstrate its improvement by evaluating it on a real robotics task of catching a balloon mid-air and two domain-adjacent sports tasks of badminton and baseball trajectory prediction [\textbf{Right}].
    \end{minipage}\par
    \vspace{-0.15cm}
    \label{fig:overview}
\end{strip}

\begin{abstract}
Learning generalizable dynamics models requires diverse, grounded training data. Existing approaches trained on internet-sourced data are constrained by the availability of high-quality samples, while the diversity of real-world data collection efforts is limited by infrastructure and operator demands. In our work, we propose collecting dynamics data through natural, embodied robot interactions. We present \algname{}, a robot quadruped that collects in-the-wild dynamics data by playing fetch. \algabbr{} first observes a ball trajectory thrown by a user. Then, it retrieves the ball and returns it, creating a naturally engaging and repeatable interaction that yields grounded observations of object dynamics, including the arc of throws, the effect of gravity, and the variability of real-world conditions. We collect FIDO data across $N=18$ users in three in-the-wild environments. A projectile dynamics model trained on this data exhibits strong transfer to unseen domain-adjacent tasks. \algabbr{} outperforms a pretrained VLM baseline by 29.1\% on shuttlecock trajectory prediction in badminton and by 26.8\% on ball trajectory prediction in baseball, with both results improving monotonically with the diversity of collected fetch data. Our learned dynamics model is also directly useful for robot planning: a robot plans through these learned dynamics to catch a balloon before it hits the ground, where the addition of \algabbr{} data improves catching performance by 20.0\%. These results suggest that in-the-wild interaction diversity is a key driver of learning generalizable physical dynamics. Additional visuals can be found at our website \href{https://fido-robot.github.io/site}{fido-robot.github.io/site}.
\end{abstract}

\begin{figure*}[t]
    \includegraphics[width=\textwidth]{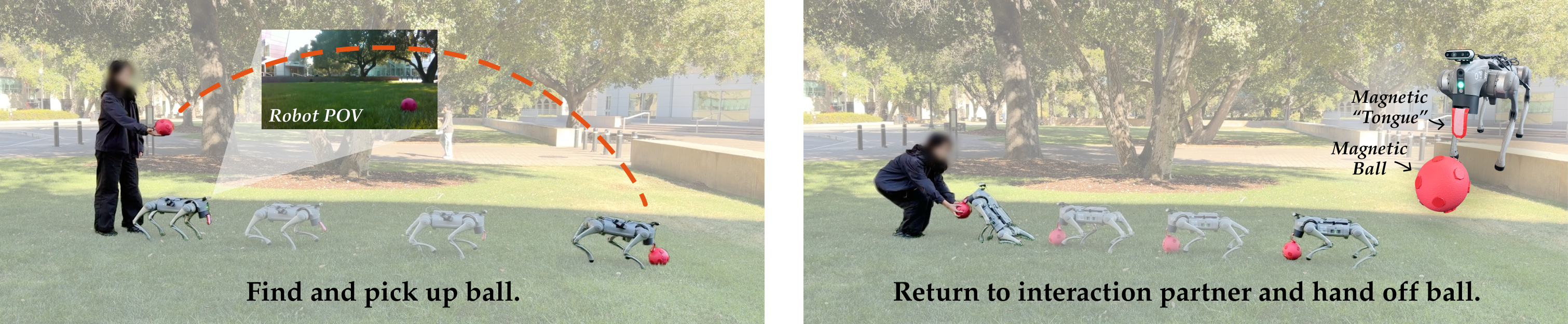}
    \caption{\textbf{In-the-Wild Data Collection.} The FIDO system is a robot quadruped that plays fetch using a magnetic ball and a felt-covered ``tongue'' (\S \ref{sec:interaction}). During this interaction, FIDO collects video data on the ball trajectory. We use these videos to fine-tune a VLM for dynamics prediction. (\S \ref{sec:adaptation}).}
    \label{fig:fetch-interaction}
\vspace*{-0.5cm}
\end{figure*}

\section{Introduction}
\label{sec:intro}

Physical reasoning is a core requirement for robots operating in the real world. Dynamics models capture representations of how objects move and interact, and are a key component of robot planning and control \cite{murray1994mathematical, qin2003survey, lelidec2024contact, collins2021review}. While traditionally limited to narrow, task-specific settings \cite{nagabandi2019ddpm, ai2025review, li2018learning}, recent advances in foundation models have opened the possibility of learning general dynamics models, capable of reasoning across tasks and physical conditions \cite{kim2026cosmos}.
Realizing this potential, however, requires observations that expose the physical events these models must learn. 

Internet-sourced training data is abundant but the relevant physical interactions occur only incidentally, making it difficult to curate data aligned with downstream robotics and domain-adjacent tasks~\cite{kim2026cosmos, zhu2024survey}. 
Collecting \textit{grounded, real-world} training data that is aligned with the physical phenomena of interest while spanning enough variation to support transfer remains an open challenge.

Prior work has explored crowdsourcing as a path to data diversity. Human participation can yield large and varied datasets, but curating a useful dataset often requires significant infrastructure, operator expertise, or explicit incentive mechanisms~\cite{mandlekar2019roboturk, mirchandani2025robocrowd}. These demands make in-the-wild collection difficult in practice. 
One source of this overhead is that crowdsourced robotics data are often framed as soliciting expert task demonstrations \cite{mirchandani2025robocrowd}. 
Unlike imitation learning, however, dynamics learning does not require every interaction to demonstrate expert or optimal behavior~\cite{sun2026latentpolicybarrierlearning}: a casual interaction still provides valid observations of how the world responds. 
\uline{We propose using natural human–robot interactions as a data-collection mechanism.} When an interaction is easy to understand, engaging, and naturally repetitive, participants generate dynamics-relevant observations as a byproduct of interacting with the robot. This produces data that is intentionally aligned with the target physical phenomenon while retaining the variation that arises across people and environments.
 
We instantiate this principle in \algabbr{} (\textbf{F}etch \textbf{I}nteraction for \textbf{D}ynamics \textbf{O}bservation), a quadruped robot that plays fetch with a human partner. During each round, the robot observes a thrown ball, retrieves it, and returns it to the participant. Fetch provides a simple experimental setting for studying whether casual interaction can yield transferable dynamics knowledge. Each round produces observations of physically meaningful events: the arc of a throw, the pull of gravity, and the bounce on landing. The resulting data captures a greater variety of physical conditions than laboratory collection allows (e.g.,\ backgrounds, lighting and weather conditions, surface types, and throwing styles). 
Unlike passive internet video, active collection from FIDO makes relevant physical events directly elicitable, rather than contingent on their availability in existing footage.
Because collection can take place outside a controlled setting and does not depend on a fixed set of trained operators, we refer to it as \textit{in-the-wild}.
\uline{Our key contribution is an intuitive system for in-the-wild data collection that turns casual human play into observations of real-world object dynamics.}

Using the FIDO system, we collect data across 18 interaction partners and three environments. We fine-tune a Vision Language Model (VLM) for dynamics prediction on this data (\S \ref{sec:adaptation}) and verify its generalizability in two experimental setups. First, we demonstrate transfer to two unseen domain-adjacent tasks: shuttlecock trajectory prediction in badminton and ball trajectory prediction in baseball. Compared to the VLM before training on FIDO data, the \algabbr{}-trained model achieves a \textbf{29.1\%} and \textbf{26.8\%} improvement on shuttlecock and baseball prediction respectively, with performance improving monotonically with data diversity (\S \ref{sec:gen_dynamics_modeling}). Second, we show that the \algabbr{}-trained model is effective on a real-world dynamics prediction task: catching a balloon in a basket held by the robot before it hits the ground (\S \ref{sec:balloon}). \algabbr{} achieves \textbf{20.0\%} better catching performance than baselines trained without \algabbr{} data. 
Together, these results show that data elicited through a simple human-robot interaction can improve dynamics prediction across unseen objects and tasks and support downstream real-world robot behavior.

\section{Related Work}
\label{sec:related_work}

\noindent \textbf{Learning Dynamics Models.} Past approaches to learning dynamics models have focused on task-specific settings, relevant to a single robot or environment \cite{nagabandi2019ddpm, finn2017deep, ai2025review, wang2023dynamic, driess2022compNerf, hafner2018planet, xu2025neural}. While effective within their target domains, their performance often degrades outside the training distribution. Large-scale foundation models have enabled more general-purpose modeling \cite{Qwen3-VL, comanici2025gemini25pushingfrontier}, with recent works demonstrating that Vision-Language Models (VLMs) can capture diverse object dynamics and can be leveraged for planning \cite{kim2026cosmos, tran2026egoexo, zhu2024survey, bardes2024revisiting}. 
However, these models are often trained on broad corpora that were not constructed to repeatedly expose a target physical event, such as object flight and impact \cite{kim2026cosmos, wan2025, sora2024}.
Our work complements these model-centric advances by studying the source of adaptation data. The FIDO system elicits repeated object-motion events through a fetch interaction. We evaluate whether finetuning a VLM on trajectory prediction on this data supports transfer across environments, viewpoints, and object-motion regimes. We also compare the FIDO-trained model with a model fine-tuned on curated internet videos of object motion.

\begin{figure}[t]
    \centering
    \includegraphics[width=0.9\columnwidth]{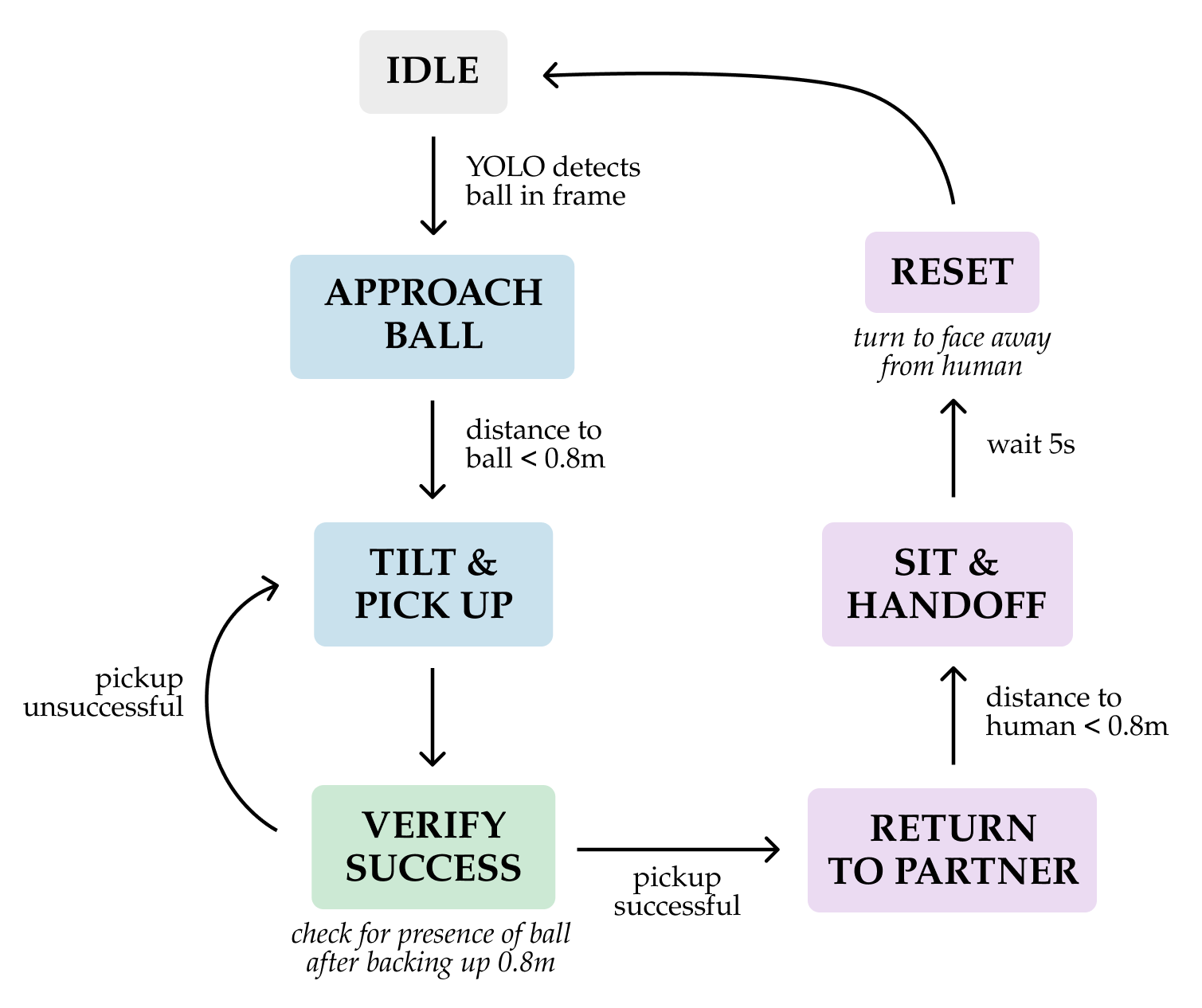}
    \caption{\textbf{State Machine of Fetch Behavior.} The \algabbr{} system is a finite state machine consisting of seven states. Each round of fetch starts in an idle state. Upon detecting a ball in the camera view, the robot moves to pick up and return the ball to the interaction partner. At the end of each interaction, the robot resets to a starting position facing away from the interaction partner and waits for the next throw.}
    \label{fig:fsm}
    \vspace*{-0.5cm}
\end{figure}

\noindent \textbf{Crowdsourcing Data.} Crowdsourcing is an appealing option for collecting large and diverse datasets across many application domains \cite{imagenet, amt2012, prolific2026, vaish2014twitch}. In robot learning, Mandlekar et al. \cite{mandlekar2019roboturk} use a mobile phone teleoperation interface to collect manipulation data remotely at scale. Mirchandani et al. \cite{mirchandani2025robocrowd} investigate incentive strategies for crowdsourcing data collection using a bimanual robot teleoperation system deployed in a public space. Other works have developed new hardware interfaces — sensorized grippers, portable motion capture systems, or egocentric wearables — to enable demonstration collection in the real world without requiring access to a physical robot \cite{liu2025egozerorobotlearningsmart, kareer2024egomimicscalingimitationlearning, chi2024universal}. These approaches broaden demonstration collection, but participants must still deliberately produce useful demonstrations through specialized hardware, interfaces, or collection protocols. FIDO changes the contribution required of the participant: rather than producing a prescribed task demonstration, the participant plays a familiar game, and dynamics observations arise from the resulting physical interaction. Because the learning target is the object motion rather than the participant’s proficiency, nonexpert variation in ball trajectory remains useful in training.

\noindent \textbf{In-the-Wild Robot Data Collection.} Deploying robots in homes, outdoor spaces, and other uncontrolled environments exposes them to physical conditions that are difficult to enumerate in the laboratory \cite{du2026snoopie, liu2024okrobot, fu2024mobile, intelligence2025pi05visionlanguageactionmodelopenworld, shafiullah2023bringing}. These deployments provide not only a test of robustness but also a source of data for continued model adaptation. UMI \cite{chi2024universal} demonstrates that collecting manipulation demonstrations in diverse, real-world environments rather than fixed lab settings enables zero-shot generalization to novel environments and objects. UMI increases accessibility by decoupling demonstration collection from the target robot. FIDO takes a complementary approach: it retains the robot but reduces participant burden by embedding data collection in a familiar interaction.

Most closely related to our work, Grannen et al. \cite{grannen2025scanford} use an in-the-wild robot platform to collect data that continually improve a VLM's Optical Character Recognition (OCR) capabilities, and similarly demonstrate transfer to domain-adjacent settings. However, their evaluation centers on a static perceptual recognition task. Our work shares Grannen et al.’s view of in-the-wild deployment as an opportunity for continued model improvement, but connects the learned model back to robot action: we target \textit{dynamics prediction} and deploy the model in a real-world robot balloon-catching task.

\section{\algabbr{}: A System for In-the-Wild Data Collection}
\label{sec:interaction}

Our method, \algname{}, considers the setting of ``fetch'', in which a user throws a ball for the robot to retrieve. The naturally repetitive and unconstrained nature of this task yields a diverse set of video observations of thrown ball trajectories, which we use downstream to learn generalizable projectile-style dynamics models.

\noindent \textbf{Hardware Setup.}
We use a Unitree Go2 quadruped robot as our data collection platform. The Go2's portability and ability to traverse varied terrain enables usage in diverse environments, ideal for in-the-wild data collection. We additionally mount an Intel RealSense D435i depth camera facing horizontally forward on the robot's head.

Since the Go2 lacks native manipulation capabilities, we design a custom end-effector to enable ball retrieval. Our mechanism consists of a magnet enclosed in a felt ``tongue", secured by a 3D-printed mount in place of the stock LiDAR unit at the base of the robot's head. We attach magnets of opposite polarity on the surface of the ball, a foam dodgeball, creating a passive retrieval mechanism that does not require additional actuation (\cref{fig:fetch-interaction} Right). The zoomorphic embodiment of the Go2 naturally affords the fetch interaction, lowering the barrier to crowdsourcing data by grounding data collection in a familiar activity that requires little training or instruction for the interaction partner.

\noindent \textbf{Fetch Interaction.}
The interaction begins with the robot standing at the interaction partner's feet and facing away from them (\cref{fig:fetch-interaction}). The interaction partner is instructed to throw the ball into view of the RealSense camera mounted on the robot head (approx.\ $70^{\circ}$ horizontal FOV). Once the robot detects the ball with an off-the-shelf object detection model, it begins approaching the ball. Upon reaching a close range, the robot tilts forward to lower the ``tongue'' end-effector and attempts to magnetically attach to the ball by moving towards the last known position of the ball. The robot then returns to a neutral stance to lift the ball off the ground and reverses to verify success of the retrieval. If the ball is still visible after backing up, the robot reattempts the approach. Once the robot has successfully obtained the ball, it navigates back towards its initial position, using an object detection model to identify the interaction partner. Finally, the robot assumes a sitting position in front of the person to hand off the ball.

\noindent \textbf{Implementation Details.}
The fetch behavior is implemented using a finite state machine (\cref{fig:fsm}) which responds to perception and proprioception feedback. We use YOLO11n~\cite{Jocher_Ultralytics_YOLO_2023} as our off-the-shelf object detector to run real-time ball and person detection onboard the robot. For each fetch interaction, we record video data from the RealSense camera and save the segment from which the ball first appears in the frame, until the robot begins to move to fetch the ball. The robot approaches the ball by running a PD controller to perform visual servoing. We control the dog's gait using a pretrained locomotion policy provided through the high-level interface in the Unitree SDK~\cite{unitree2026sports}.

\begin{figure*}[t]
    \centering
    \includegraphics[width=\textwidth]{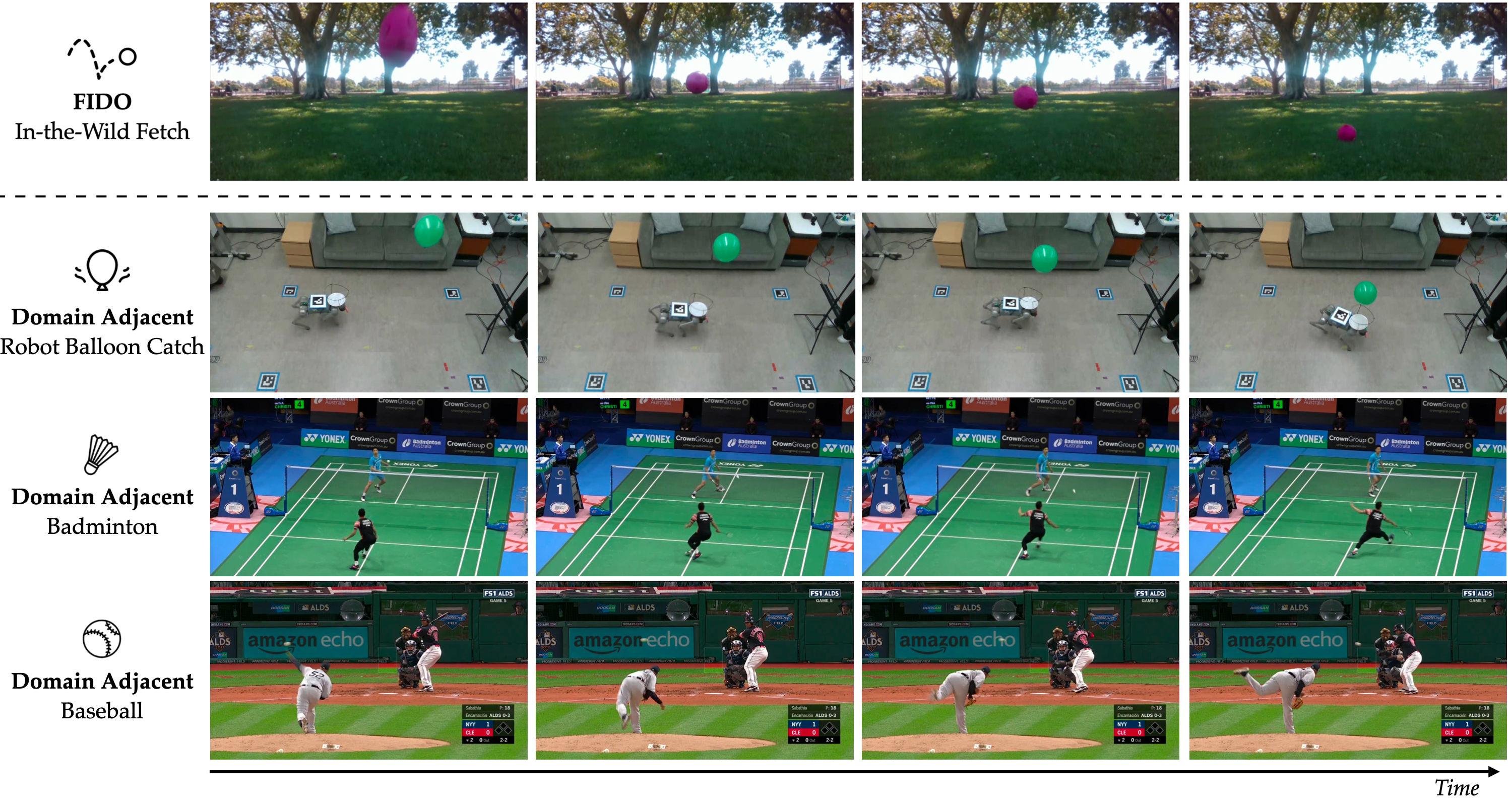}
    \caption{\textbf{FIDO Data and Domain Adjacent Test Settings.} We evaluate the generalizability of dynamics models trained on \algabbr{} data on three different settings. The domain-adjacent domains of Balloon Catch, Badminton, and Baseball differ substantially from the FIDO data in object appearance, scale, and aerodynamic properties, as well as in camera viewpoint and environmental context. Our goal is zero-shot transfer to the sport domains without any in-domain training data, testing whether grounded in-the-wild fetch interactions can yield dynamics models that generalize across this diversity.}
    \label{fig:domains}
\end{figure*}

\section{Learning a Dynamics Model with \algabbr{}}
\label{sec:adaptation}
We use videos obtained from \algabbr{}'s fetch interactions to adapt a dynamics model that predicts future object locations given past locations. In this section, we first formulate the dynamics prediction problem, then we describe how we obtain necessary supervision from \algabbr{}, and finally explain how we adapt a pretrained VLM using this supervision. 

\noindent \textbf{Problem Setting.}
We model object dynamics as future prediction in image space. At time $t$, the dynamics model $f_\theta$ observes a past image $I_{t - h + 1}$, and a short history of object locations $H_{t} = \{ p_{t - h + 1}, \ldots, p_{t}\}$, where each location is represented by its pixel position in the image $p_{t} = (x_t, y_t)$. The model is further provided with a query specifying which future timesteps to predict: $\Delta = \{\delta_1, \delta_2, \ldots, \delta_{K} \}$. The dynamics model is expected to predict the corresponding set of future object positions $Y_t = \{p_{t+\delta_1}, \ldots, p_{t + \delta_K}\}$. We write the dynamics model as: $\hat{Y}_t = f_\theta(I_{t - h + 1}, H_t, \Delta)$. Here, $I_{t - h + 1}$ provides visual context at the beginning of the trajectory, while the history $H_t$ provides recent motion of the object. The offset query allows the model to be transferred to varied use cases without changing model architecture. For example, catching a falling balloon requires long-horizon predictions while predicting whether a shuttlecock will land in bounds might require shorter-horizon predictions.

\noindent \textbf{Trajectory Supervision with \algabbr{}.}
 We learn a dynamics model using trajectory supervision extracted from \algabbr{} videos, broadly focusing on modeling the ``throwing" motion. Each fetch interaction provides a video sequence $V=\{I_1, \ldots, I_T\}$ demonstrating motion of a ball thrown by an interaction partner. We use state-of-the-art object detection and tracking methods~\cite{medeiros2023langsegmentanything, harley2025alltracker} to obtain the trajectory of the ball in image space, $\{p_1, p_2, \ldots p_T\}$, where $p_t$ is the tracked pixel location of ball center. We sample an observation at time $t$ such that a trajectory contains a valid history window and enough timesteps into the future to construct a prediction target. For each sampled observation, we randomly choose the number and stride of future predictions, while ensuring that all requested future timesteps are available in the trajectory. This generates multiple training examples by varying the observation and the future offset query from a single trajectory. In our implementation the history contains five observed points, which we found in practice to contain the necessary information for predicting future trajectories, such as the velocity and initial position of the ball. %

\noindent \textbf{VLM Adaptation to Dynamics Prediction.}
 Our goal with \algabbr{} is to build a generalizable dynamics model, enabling transfer to related projectile trajectory prediction settings. Hence, we instantiate the dynamics model as a VLM which incorporates priors from large-scale Internet data. While state-of-the-art VLMs can produce qualitatively plausible outputs on physical reasoning tasks, their answers are often numerically inaccurate \cite{li2025quantiphy}. In this work, we improve a pretrained VLM's ability to predict physically realistic trajectories using \algabbr{} data. To adapt the pretrained model efficiently while retaining its broad visual priors, we train it using low-rank adaptation (LoRA)~\cite{hu2022lora}, which keeps the pretrained backbone fixed and learns a small set of trainable low-rank updates to selected weight matrices.

 \begin{figure}[h]
    \centering
    \vspace{-0.3cm}
    \includegraphics[width=\columnwidth]{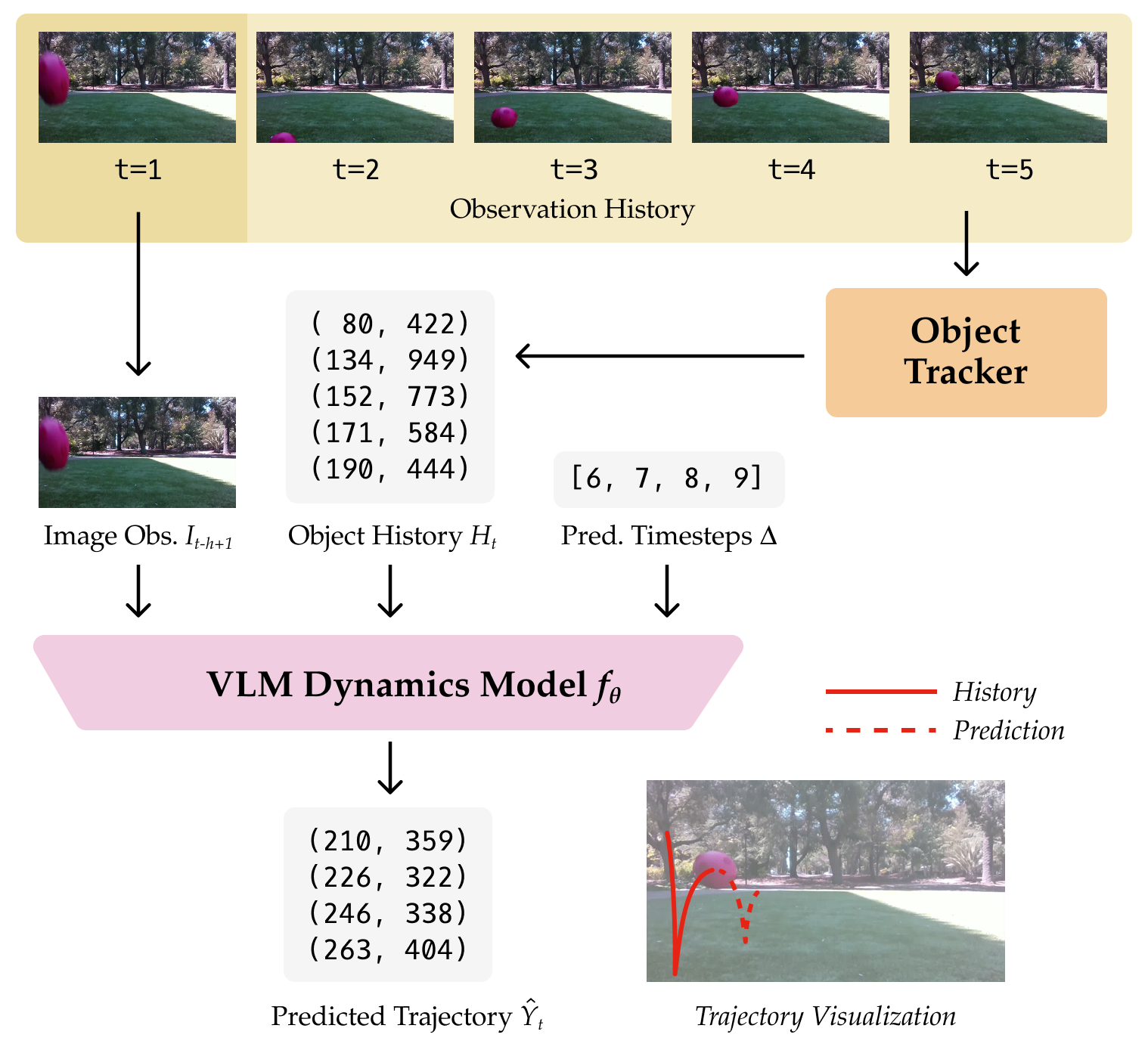}
    \caption{\textbf{VLM Dynamics Model.} We fine-tune a pretrained VLM on \algabbr{} data to learn an image-space dynamics model. We extract object trajectories from video using state-of-the-art object detection and tracking methods. From image input $I_{t-h+1}$ and object location history $H_t$, the VLM $f_\theta$ predicts future object locations $\hat{Y}$ at specified timesteps $\Delta$.}
    \vspace{-0.3cm}
\end{figure}

\section{Experimental Results}
\label{sec:experiments}

We evaluate our \algabbr{}-trained dynamics model on three tasks, as shown in \cref{fig:domains}. First, we test the generalizability of the learned dynamics across two unseen domain-adjacent tasks: shuttlecock trajectory prediction in badminton and ball trajectory prediction in baseball. Second, we demonstrate its utility for robot planning on a real robotics task of catching a falling balloon before it hits the ground. Across both settings, we investigate whether the diversity of in-the-wild fetch data is the key driver of generalization.

\begin{table*}[t]
\centering
\setlength{\tabcolsep}{9pt}
\setlength{\aboverulesep}{0.2ex}
\setlength{\belowrulesep}{0.2ex}
\begin{tabular}{l  c  c @{\hspace{7pt}}|@{\hspace{7pt}} c  c  @{\hspace{7pt}}|@{\hspace{7pt}} c c }
\toprule
\multirow{2}{*}[0.9ex]{\thead{Model}} & \multirow{2}{*}[0.9ex]{\thead{Num. \\ Throwers}} & \multirow{2}{*}[0.9ex]{\thead{Num. \\ Envs.}} & \multicolumn{2}{c @{\hspace{7pt}}|@{\hspace{7pt}}}{\textbf{Badminton Shuttlecock } \raisebox{-0.5mm}{\includegraphics[height=4mm]{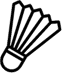}}}  & \multicolumn{2}{c}{\textbf{Baseball Ball } \raisebox{-0.5mm}{\includegraphics[height=4mm]{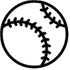}}} \\ %
& & & RMSE ($\downarrow$) & DTW ($\downarrow$) & RMSE ($\downarrow$) & DTW ($\downarrow$) \\
\noalign{\smallskip}
\toprule
\noalign{\smallskip}

      Pretrain VLM & -- & -- & $0.55 \pm 0.08$ & $2.50 \pm 0.37$ & $0.41 \pm 0.06$ & $3.17 \pm 0.48$ \\
  Classical predictor (Kalman) & -- & -- & $0.79 \pm 0.02$ & $4.35 \pm 0.16$ & $0.59 \pm 0.01$ & $5.19 \pm 0.15$ \\
  Coordinates-only LLM & -- & -- & $0.54 \pm 0.05$ & $2.83 \pm 0.27$ & \textbf{0.30 $\pm$ 0.03} & \textbf{2.42 $\pm$ 0.28} \\
  Internet-tuned VLM & -- & -- & $0.52 \pm 0.07$ & $2.41 \pm 0.37$ & $0.38 \pm 0.05$ & $3.01 \pm 0.40$ \\ %
  \midrule
  \multirow{3}{*}{\algabbr{}-tuned VLM} & 2 & 1 & $0.59 \pm 0.04$ & $3.04 \pm 0.26$ & $0.63 \pm 0.02$ & $5.59 \pm 0.17$ \\
   & 2 & 3 & $0.50 \pm 0.05$ & $2.53 \pm 0.28$ & $0.51 \pm 0.03$ & $4.39 \pm 0.31$ \\   %
   & 18 & 3 & \textbf{0.39 $\pm$ 0.05} & \textbf{1.87 $\pm$ 0.26} & \textbf{0.30 $\pm$ 0.03} & \textbf{2.35 $\pm$ 0.25} \\

\bottomrule
\end{tabular}
\vspace{1.0em}

\refstepcounter{table} %
\begin{minipage}{1.0\textwidth}
    \footnotesize
    \setlength{\parskip}{0pt} %
    \setlength{\parindent}{0pt} %
    Table~\thetable. 
    \textbf{\algabbr{}'s in-the-wild data collection yields more diverse data, leading to better domain-adjacent transfer.} We evaluate the generalizability of dynamics models on two unseen trajectory prediction domains: badminton shuttlecock~\cite{sun2020tracknetv2} and baseball~\cite{piergiovanni2018fine}. We report the average and 95\% confidence intervals of root mean squared error (RMSE) and dynamic time warping (DTW), both normalized by trajectory geodesic distance in pixels. 
    Fine-tuning on the most diverse set of fetch data across throwers and environments achieves the best performance across both metrics and both domains.
\end{minipage}
\label{table:sports}
\vspace{-1.2em}

\end{table*}
\subsection{\algabbr{} Dataset}
The \algabbr{} dataset consists of 75 videos collected across three environments with 18 interaction partners, solicited as passersby in public outdoor areas. Each video captures a single throw of the ball, and the dataset is gathered over 59 minutes of human interaction time.

Following collection, we post-process the video data to isolate the trajectory of the ball for each throw. Each trajectory segment begins with the first video frame where the ball is visible and ends with the last frame before the robot moves to retrieve the ball. The average trajectory clip length per throw after post-processing is 1.64 seconds.

\subsection{Generalizable Dynamics Modeling}
\label{sec:gen_dynamics_modeling}
We consider two domain-adjacent trajectory prediction tasks to evaluate the generalizability of our fetch-data trained dynamics model as visualized in \cref{fig:domains}. The first is shuttlecock trajectory prediction in badminton, evaluated on the TrackNet dataset~\cite{sun2020tracknetv2}. The second is ball trajectory prediction in baseball, evaluated on the MLB-YouTube dataset~\cite{piergiovanni2018fine}. Both tasks share underlying physical structure with fetch, specifically projectile motion under gravity, but differ in object properties, camera perspectives, and environmental context, making them a meaningful test of out-of-domain generalization.

For badminton, we evaluate on $100$ TrackNetV2 trajectory segments constructed from official shuttlecock annotations. For baseball, we evaluate on $77$ pitch trajectories from MLB-Youtube. Since MLB-Youtube does not provide frame-level ball locations, we initialize ball detections with LangSAM~\cite{medeiros2023langsegmentanything} and track them across the clip using AllTracker~\cite{harley2025alltracker}, followed by automatic filtering and manual removal of failed tracks.

We fine-tune Qwen3-VL-2B-Instruct~\cite{Qwen3-VL} on fetch data collected by FIDO and evaluate the resulting dynamics model on both tasks. 
We use LoRA finetuning with LoRA rank $8$, $\alpha=16$ and dropout $0.05$, and optimize with AdamW at a learning rate of $10^{-5}$ with cosine decay. We train with an effective batch size of 8 for 1000 gradient steps. 

\noindent \textbf{Baselines.} We compare a \algabbr{}-trained model against four baselines: 1) the pretrained VLM, 2) a classical trajectory predictor tuned on \algabbr{} trajectories, 3) a coordinates-only LLM finetuned on \algabbr{} trajectories, and 4) a VLM fine-tuned using the FIDO method with object trajectory videos from the internet instead of FIDO data. 

For the classical baseline (\#2), we tune a family of classical trajectory predictors using FIDO trajectories, including constant-velocity and constant-acceleration Kalman filters, least-squares linear and quadratic trajectory fits, particle filters, and variants that blend their predictions with the last observed position. We sweep $14,056$  configurations over predictor type, process and observation noise, acceleration damping, and the weighting of the last-position prediction, and select the lowest RMSE variant.
For the coordinates-only baseline (\#3), we test whether the image adds useful information beyond the observed trajectory. We therefore finetune the same model on the same FIDO trajectories while removing the image input, leaving only the trajectory history $H_t$ and timestep query $\Delta$ as inputs.
Next, to test whether similar supervision can be obtained from readily available video rather than robot-collected interactions, we fine-tune the same VLM on projectile trajectories extracted from internet videos (\#4). We collect candidate projectile-motion clips from UCF-101~\cite{soomro2012ucf101dataset101human} and process them using the same trajectory extraction pipeline applied to the MLB dataset, discarding clips with tracking failures or unsuitable motion. From the remaining clips, we sample $75$ trajectories to match the size of the \algabbr{} dataset.

\noindent \textbf{Diversity Ablations.} Finally, to investigate the role of data diversity in generalization, we conduct ablations by fine-tuning on equally-sized datasets of fetch data with varying diversity: 
\begin{enumerate}
  \item Fine-tuning with fetch data collected in \textbf{1 environment} with \textbf{2 interaction partners},
  \item Fine-tuning with fetch data collected across \textbf{3 environments} with the same \textbf{2 partners},
  \item Fine-tuning with \algabbr{} data collected \textit{in-the-wild} across \textbf{3 environments} and with \textbf{18 interaction partners}.
\end{enumerate}
The datasets for (1) and (2) are disjoint from the original \algabbr{} dataset and are collected with two authors serving as interaction partners. By comparing performance across these conditions, we aim to demonstrate that diversity in data collection, both across interaction and environment, is a key driver of transfer to unseen domains.

\noindent \textbf{Results.} As shown in \cref{table:sports}, we measure the success of our generalizability with two metrics: root mean squared error (RMSE) and dynamic time warping (DTW), both normalized to the ground-truth length of the predicted trajectory. We report normalized RMSE and normalized DTW averaged across evaluation trials and three independent training seeds. 
The reported intervals denote $95\%$ hierarchical bootstrap confidence intervals over evaluation trajectories and training seeds. For fixed baselines (first two rows in~\cref{table:sports}) that are not trained in our experiments, intervals reflect evaluation-trajectory variability only.
\uline{The VLM tuned on the full FIDO set achieves the best RMSE and DTW performance for badminton shuttlecock trajectory prediction, outperforming all baselines and lower-diversity fetch data ablations.} We observe consistent improvement in both metrics with increasing data diversity, from a single environment and two throwers, to three environments, to the full in-the-wild collection. Overall, \algabbr{} achieves a 29.1\% improvement in RMSE and a 25.2\% improvement in DTW over the pretrained baseline, and outperforms the best lower-diversity ablation by 22.0\% in RMSE and 26.1\% in DTW.

We observe a similar trend in the baseball setting, where model performance on both RMSE and DTW improves monotonically with the diversity of fetch data collected. However, unlike badminton, the pretrained baseline outperforms both lower-diversity fetch data ablations. We hypothesize that this is because baseball ball dynamics are far more common in internet-scale training data than badminton shuttlecock dynamics, giving the pretrained model a stronger prior for this domain. Fine-tuning on fetch data consisting of relatively homogeneous throw styles (from only two throwers) appears to overfit to fetch-specific trajectory patterns, degrading generalization to baseball dynamics. With sufficient diversity, however, this effect is overcome -- a model fine-tuned with in-the-wild \algabbr{} data outperforms the pretrained baseline by 26.8\% and 25.9\% across RMSE and DTW respectively.

Together, these results highlight the need for diverse collection conditions to enable generalizable learnings, even in domains well-represented in pretraining data. Critically, this diversity is made possible by \algabbr{}'s data collection through natural interaction. By lowering the barrier to participation, \algabbr{} naturally accumulates a breadth of throwing styles and environments, driving generalization.

\begin{table*}[t!]
\centering
\setlength{\aboverulesep}{0.2ex}
\setlength{\belowrulesep}{0.2ex}
\begin{minipage}{0.75\textwidth}
\centering
\setlength{\tabcolsep}{4pt}
\begin{tabular}{l !{\hspace{2pt}}|!{\hspace{5pt}} c c c c !{\hspace{5pt}}|!{\hspace{5pt}} c c}
\toprule
\multirow{2}{*}[0.9ex]{\thead[l]{Fine-Tune\\Dataset}}
& \multicolumn{4}{c !{\hspace{5pt}}|!{\hspace{5pt}}}{\textbf{Robot Balloon Catching } }
& \multicolumn{2}{c}{\textbf{Balloon Dynamics Modeling} \raisebox{-0.8mm}{\includegraphics[height=4mm]{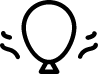}}} \\
 & Caught & Rim & Miss & Score & RMSE ($\downarrow$) & DTW ($\downarrow$) \\
\noalign{\smallskip}
\toprule
\noalign{\smallskip}
Pretrain & 12/50 & 27/50 & 11/50 & 25.5 & $1.28 \pm 0.69$ & $5.16 \pm 1.91$ \\
\algabbr{} & 12/50 & 29/50 & 9/50 & 26.5 & $0.85 \pm 0.26$ & $3.35 \pm 0.83$ \\
Few-Shot Balloon & 12/50 & 31/50 & 7/50 & 27.5 & $0.81 \pm 0.27$ & $3.19 \pm 0.97$ \\
\algabbr{} + Few-Shot Balloon & \textbf{17/50} & \textbf{32/50} & \textbf{1/50} & \textbf{33.0} & \textbf{0.78 $\pm$ 0.27} & \textbf{3.11 $\pm$ 0.93} \\
\bottomrule
\end{tabular}
\end{minipage}
\hfill
\begin{minipage}{0.2\textwidth}
\includegraphics[width=0.8\linewidth]{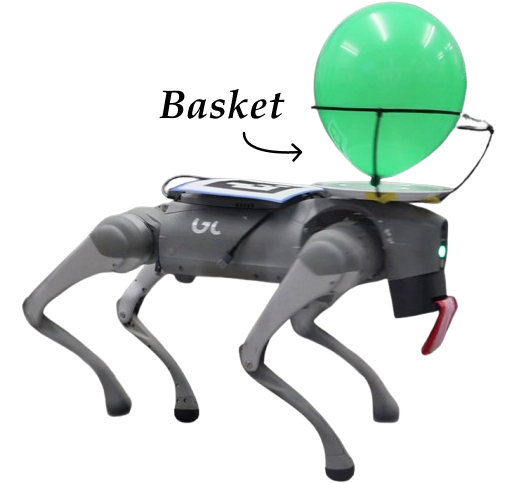}
\end{minipage}
\vspace{0.5em}
\refstepcounter{table} %
\begin{minipage}{1.0\textwidth}
    \footnotesize
    \setlength{\parskip}{0pt} %
    \setlength{\parindent}{0pt} %
    Table~\thetable. 
    \textbf{Planning and offline evaluation with a \algabbr{} Dynamics Model for Balloon Catch [\textbf{Right}].} We evaluate dynamics models fine-tuned on different data mixtures for the task of physical robot balloon catching (Caught / Rim / Miss) and on offline balloon dynamics modeling (RMSE / DTW). We consider catching success classified as either a successful catch, a rim hit, or a miss. We also calculate a composite score \texttt{\#Caught} + $0.5 \times$\texttt{\#Rim}. Combining \algabbr{} data with a small amount of balloon-specific data yields the strongest performance across both the physical catching task and offline trajectory prediction. Notably, this combined model outperforms both the pretrained baseline and the balloon-only fine-tuned model, suggesting that the diversity of \algabbr{}'s fetch data provides a meaningful foundation for downstream robot planning.
\end{minipage}
\label{table:balloon}
\vspace{-1.7em}
\end{table*}

\subsection{Robot Balloon Catching} 
\label{sec:balloon}

We consider the task of catching a thrown balloon in a basket attached to the head of the robot. This task tests both the generalization capabilities of \algabbr{} to a novel dynamics situation and its applicability to a downstream robot task. Because the robot has a limited speed, it strongly benefits from proactively moving to where the balloon will land, which is predicted by our dynamics model. Thus, the catching success is a reflection of the performance of the dynamics model predictions. 

We employ an angled, overhead Intel RealSense D435i camera with the full arena workspace in view, as demarcated by four Aruco tags. Similar to \cref{sec:interaction}, the position of the balloon in the scene is tracked using YOLO11n~\cite{Jocher_Ultralytics_YOLO_2023}. A context of past balloon positions, as well as an initial image of the scene, are fed into the Qwen3-VL-2B-Instruct~\cite{Qwen3-VL} dynamics model to predict the future trajectory of the balloon. Given these dynamics predictions, we leverage known camera extrinsics and an Aruco tag mounted on the back of the robot to determine the waypoints for the robot to execute, computed as a linear trajectory towards the predicted future location of the balloon.  

We evaluate performance across four VLM variants with increasing in-domain data exposure:
\begin{enumerate}
    \item \textbf{Pretrain}, which measures out-of-the-box performance of the VLM without finetuning on balloon or FIDO data,
    \item \textbf{FIDO}, which measures performance after finetuning with 75 videos of \algabbr{} data,
    \item \textbf{Few-Shot Balloon}, which measures performance after fine-tuning with five in-domain balloon throw trajectories,
    \item \textbf{FIDO + Few-Shot Balloon}, which measures performance after fine-tuning with a combination of FIDO data and five in-domain balloon throw trajectories.
\end{enumerate}

Due to the inference latency of the VLM (650ms on an L40S GPU), we fill our balloon with a mixture of helium and air to increase its buoyancy and slow its fall. We conduct $50$ trials of balloon catch for each method and record three outcomes: \texttt{Caught}, when the balloon touches the bottom of the basket; \texttt{Rim}, when the balloon hits the rim of the basket before falling out; and \texttt{Miss}, when the balloon fails to make contact with any part of the basket. We report outcome counts and a composite score \mbox{\texttt{\#Caught} + $0.5 \times$\texttt{\#Rim}}.

\noindent \textbf{Results.} As shown in \cref{table:balloon}, fine-tuning on \algabbr{} fetch data alone yields modest improvements over the pretrained baseline in balloon catching, reducing misses from 11/50 to 9/50. We attribute the limited gains to the domain gap between fetch and balloon dynamics. While both involve falling objects, balloons are significantly more affected by air resistance and drift, making direct transfer from fetch data alone insufficient. Fine-tuning on a small amount of balloon-specific data similarly improves over the pretrained baseline (7/50 misses), but without sufficient diversity in the training distribution, the model struggles to robustly generalize across the range of balloon trajectories encountered at test time. Combining \algabbr{} fetch data with the few-shot balloon data results in the strongest performance, catching 17/50 balloons and nearly eliminating misses (1/50), achieving the highest composite score across conditions. 

We additionally perform offline evaluations on the 200 balloon trajectories recorded from the trials above. Because the VLM has nontrivial inference latency, these results isolate the quality of dynamics predictions from physical task execution. The offline results align with performance in balloon catching, where fine-tuning on a combination of \algabbr{} and few-shot balloon data yields the best dynamics modeling.
We believe the diverse fetch data provides a strong foundation for the general dynamics of falling objects under gravity, which the balloon-specific data then refines to align with this narrower domain. This complementary combination is enabled by \algabbr{}'s in-the-wild data collection strategy, which provides the breadth of physical conditions necessary to serve as a generalizable prior for downstream dynamics tasks.

\section{Discussion and Conclusion}
\label{sec:discussion}

In this work, we present \algabbr{}, a robot quadruped that collects grounded, in-the-wild dynamics data by playing fetch. By designing an interaction that is natural, repeatable, and accessible to non-expert participants, we enable data collection across diverse environments and a broad pool of interaction partners without specialized training or infrastructure. \algabbr{} generalizes to two unseen domain-adjacent tasks of badminton shuttlecock and baseball ball trajectory prediction with performance improving monotonically with data diversity. Compared to the pretrained baseline, \algabbr{} achieves a \textbf{29.1\%} and \textbf{26.8\%} improvement on shuttlecock trajectory prediction in badminton and ball trajectory prediction in baseball respectively. Additionally, results on both sports tasks show monotonically increasing performance with the diversity of collected data. \algabbr{}'s learned dynamics model is also directly useful for robot planning, enabling real-time balloon catching \textbf{20.0\%} better than baselines trained without \algabbr{} data. Together, these results suggest that the diversity afforded by in-the-wild data collection is central to generalizable physical dynamics learning.

\noindent \textbf{Limitations and Future Work.}
While our results with $N=18$ interaction partners across three environments are encouraging, scaling up data collection to more throwers, more environments, and longer deployment horizons would further stress-test the relationship between diversity and generalization. Additionally, our current data collection is limited to successful fetch interactions; incorporating failure modes such as missed catches and mid-air deflections, other multimodal interactions, and passive environmental observation could expose the model to a richer and more varied distribution of physical phenomena. We believe scaling up data collection could further unlock downstream applications such as reward learning and general-purpose video-based world modeling.

As data collection scales, protecting participant privacy becomes an increasingly important consideration. The \algabbr{} system captures forward-facing video, which excludes the interaction partner in the collected ball trajectories. Additional safeguards could include careful handling of incidentally captured bystanders through automated data anonymization.

\section*{Acknowledgements}
This study was determined to be exempt from review by the Stanford University Institutional Review Board.

\textit{AI Tool Disclosure}: The authors are responsible for all content in this article. AI tools (Claude, ChatGPT) were used for coding assistance.

\bibliographystyle{IEEEtran}
\bibliography{references}

\end{document}